\documentclass[runningheads]{llncs}

\usepackage{eccv}

\usepackage{eccvabbrv}

\usepackage{graphicx}
\usepackage{booktabs}
\usepackage{multirow}
\usepackage{makecell}

\usepackage[accsupp]{axessibility}  
\usepackage{soul}

\usepackage{hyperref}

\usepackage{orcidlink}

\usepackage{nicematrix} 
\usepackage{multirow} 
\usepackage{booktabs} 
\usepackage{amsmath} 
\usepackage{diagbox} 
\usepackage{array}
\usepackage{float}

\usepackage{arydshln}

\newcommand{\thickhline}{%
    \noalign {\ifnum 0=`}\fi \hrule height 1pt
    \futurelet \reserved@a \@xhline
}

\usepackage{amssymb}
\usepackage{pifont}

\begin{document}

\title{CMTA: Cross-Modal Temporal Alignment for Event-guided Video Deblurring} 

\newcommand\CoAuthorMark{\footnotemark[\arabic{footnote}]}
\author{Taewoo Kim\orcidlink{0000-0002-8608-9514}\thanks{Equal contribution.} \and
Hoonhee Cho\orcidlink{0000-0003-0896-6793}\protect\CoAuthorMark \and
Kuk-Jin Yoon\orcidlink{0000-0002-1634-2756}}

\titlerunning{Cross-Modal Temporal Alignment for Event-guided Video Deblurring}

\authorrunning{Kim et al.}

\institute{Korea Advanced Institute of Science and Technology\\
\email{\{intelpro, gnsgnsgml, kjyoon\}@kaist.ac.kr}\\
}

\maketitle

\begin{abstract}
Video deblurring aims to enhance the quality of restored results in motion-blurred videos by effectively gathering information from adjacent video frames to compensate for the insufficient data in a single blurred frame.
However, when faced with consecutively severe motion blur situations, frame-based video deblurring methods often fail to find accurate temporal correspondence among neighboring video frames, leading to diminished performance.
To address this limitation, we aim to solve the video deblurring task by leveraging an event camera with micro-second temporal resolution.
To fully exploit the dense temporal resolution of the event camera, we propose two modules:
1) Intra-frame feature enhancement operates within the exposure time of a single blurred frame, iteratively enhancing cross-modality features in a recurrent manner to better utilize the rich temporal information of events,
2) Inter-frame temporal feature alignment gathers valuable long-range temporal information to target frames, aggregating sharp features leveraging the advantages of the events.
In addition, we present a novel dataset composed of real-world blurred RGB videos, corresponding sharp videos, and event data. 
This dataset serves as a valuable resource for evaluating event-guided deblurring methods. 
We demonstrate that our proposed methods outperform state-of-the-art frame-based and event-based motion deblurring methods through extensive experiments conducted on both synthetic and real-world deblurring datasets. The code and dataset are available at \url{https://github.com/intelpro/CMTA}.
\keywords{Temporal alignment \and Video Deblurring \and Event cameras}
\end{abstract}

\section{Introduction}
\label{sec:intro}
Motion blur is a common artifact caused by dynamic movements within a scene or camera motion during exposure. Motion deblurring, which aims to reverse the blurring process, presents significant challenges due to variations in blur intensity influenced by scene structure and depth. To achieve high-quality deblurring, video deblurring has emerged, leveraging information from neighboring frames instead of relying solely on a single blurred image. However, identifying temporal correspondence between blurred video frames becomes challenging with extreme motion blur, hindering the extraction of valuable information from adjacent frames and impeding performance improvement.

Event cameras~\cite{gallego2020event}, with their extremely low latency (on the order of microseconds), can offer high-quality guidance for motion deblurring due to their ability to capture high-temporal resolution of brightness change.
To effectively utilize the advantages of the events, several event-guided motion deblurring works~\cite{kim2022event,sun2022event,zhang2023generalizing} have been introduced.
While these works have typically explored cross-modal feature fusion methods across different modalities, there has been limited work on leveraging the abundant temporal information in videos.

To obtain high-quality results, we emphasize the event camera's temporal continuity, focusing on its interaction with video frames that exhibit long-term temporal dependencies. 
Unlike previous event-guided motion deblurring works~\cite{zhang2023generalizing, sun2022event, Xu_2021_ICCV}, which relied on a single blurry image and corresponding events on its exposure time, we further design precise temporal feature alignment methods between neighboring video frames by leveraging the advantages of event data.
Specifically, we propose novel modules from two perspectives: intra-frame (interaction between events and frames within the exposure time) and inter-frame (interaction between different frames) perspectives.

From an intra-frame perspective within exposure time, we propose a Cross-modal Recurrent Intra-frame Feature Enhancement (CRIFE) module to better leverage the rich temporal information of the events by mutually interacting the blur frame and event features within the duration of exposure time. In this module, we perform recurrent attention-based feature enhancement using a transformer~\cite{vaswani2017attention} that better captures long-range pixel dependencies.

In temporal feature alignment with the second perspective, we propose a novel event-guided temporal feature alignment module, effectively leveraging rich temporal characteristics of the events.
Conventional frame-based temporal feature alignment methods rely on optical flow~\cite{ranjan2017optical} or deformable convolution~\cite{zhu2019deformable,chan2021basicvsr++}. 
While optical flow and deformable convolutions aid in achieving temporal feature alignment, the high computational complexity of these operations makes it challenging to execute them at a higher spatial scale.
Therefore, video frame alignment, generally conducted at lower spatial scales, leads to sub-optimal deblurring results due to the lack of spatial contexts of features.
To overcome these limitations, our temporal feature alignment module avoids relying on optical flow or deformable convolution by effectively leveraging the temporally dense advantages of the events.
Therefore, we can effectively perform temporal feature alignment across multiple visual scales as we do not rely on these complex operations but rather efficiently leverage the temporal information from the events. 
As a result, our temporal feature alignment module demonstrates a significant performance improvement in event-guided video deblurring tasks.
Since we have introduced a pioneering method for aligning temporal features using the advantages of the events, it is expected to be effectively applicable to various event-guided video restoration tasks (\eg,~event-guided video super-resolution).

Finally, we propose a novel video deblurring dataset, the EVRB dataset, composed of high-quality RGB and event data. It consists of real-world blurry videos generated by extreme motion and corresponding sharp videos for generalized applications in real-world scenarios.
The network trained on the EVRB dataset can be directly applicable to real-world scenarios, making it a valuable resource for event-guided deblurring research.

\section{Related Works}
\subsection{Video Deblurring}
Early works~\cite{aittala2018burst,su2017deep} employed CNNs that take the concatenation of adjacent frames as input to address video deblurring.  Subsequently, to better leverage temporal information, approaches have emerged that employ 3D convolutions~\cite{zhang2018adversarial,Pan_2023_CVPR}, temporal alignment modules with deformable convolutions~\cite{Wang_2019_CVPR_Workshops} and optical flow~\cite{kim2018spatio,zhang2022spatio,pan2020cascaded,wang2022MMP}.
As an alternative video alignment method, some works utilize RNN~\cite{hyun2017online,wieschollek2017learning,zhong2020efficient} and transformer~\cite{liang2022recurrent,lin2022flow} structures to propagate information from long-range video frames.
However, as the intensity of motion blur in the video increases, frame-based video alignment methods struggle to achieve accurate video alignment, leading to sub-optimal deblurring results.

\subsection{Event-guided Motion Deblurring}
An event camera can effectively be used for motion deblurring as it records motion information corresponding to the brightness differences with high temporal resolution. Efforts to utilize event cameras for motion deblurring~\cite{lin2020learning, sun2022event, sun2023event, kim2022event, Xu_2021_ICCV,zhang2021fine,cho2023non, zhang2023generalizing, zhang2022unifying, pan2019bringing, jiang2020learning} have been actively ongoing.
Recent studies have focused on effectively fusing event and image features of different modalities. 
To this end, Sun~\etal~\cite{sun2022event} employed a transformer architecture for feature fusion. 
Zhang~\etal~\cite{zhang2023generalizing} effectively combined modalities using a multi-scale architecture. 
In addition, there are also attempts to address motion deblurring by assuming challenging and general scenarios~\cite{cho2023non, kim2022event, Shang_2021_ICCV}. 
However, these studies have primarily focused on single image deblurring and do not exploit the long-range temporal information demonstrated in previous video deblurring tasks. 
To effectively acquire information that may be missing from sparse events, we introduce a novel method for accurate temporal alignment with events, utilizing information from surrounding adjacent frames in video deblurring.

\section{Event-based Video Deblurring Dataset for Real-world Blur}
\label{sec:ref_data}
\subsection{Limitation of Synthetic Blur Dataset}
Typically, the blurred images synthesis procedure adheres to the methodology outlined in~\cite{nah2017deep, su2017deep, oh2022demfi, shen2020video,nah2019ntire}, involving the averaging of consecutive video frames within a fixed-size window.
However, as recent studies~\cite{zhong2021towards, zhong2023blur, rim2020real} have discussed, blur synthesis based on discrete signals may result in shutter artifacts even when averaging high-frame-rate videos. 
Furthermore, simply averaging sharp images disregards essential elements in the blurred images such as pixel saturation~\cite{rim2022realistic}, limitations due to dynamic range, and physical noise~\cite{wei2020physics, zhang2021rethinking} in data acquired during exposure time. 
Furthermore, in the case of the event-guided motion deblurring research, existing works have enhanced the synthetic aspects by using event simulator~\cite{rebecq2018esim}.
To address these limitations, recent works have attempted to acquire real-world blur datasets using hybrid camera systems~\cite{zhong2023blur,cao2022learning,rim2020real,zhong2020efficient,zhong2021towards}.
Meanwhile, to generalize event-based motion deblurring, Sun~\etal.~\cite{sun2022event} introduced the ReBlur dataset acquired in a high-precision optical laboratory using an electronic-controlled slide-rail system. 
However, despite these successes, the ReBlur dataset has limitations due to its indoor setting, a lack of dynamic objects, and the absence of high-quality RGB data as it is acquired using the DAVIS sensor~\cite{brandli2014240}. Additionally, the sequences are relatively short(minimum six blurred frames in the sequence), making it challenging for use in event-guided video deblurring.
We have acquired a new EVRB dataset for evaluating event-guided image and video deblurring methods in real-world blurry videos. 
The EVRB dataset was captured using a customized hybrid system consisting of two RGB cameras and one event camera. This system encompasses a range of blur magnitude from slight to extreme in various urban environments.

\begin{figure*}[t]
    \centering
    \includegraphics[width=.95\linewidth]{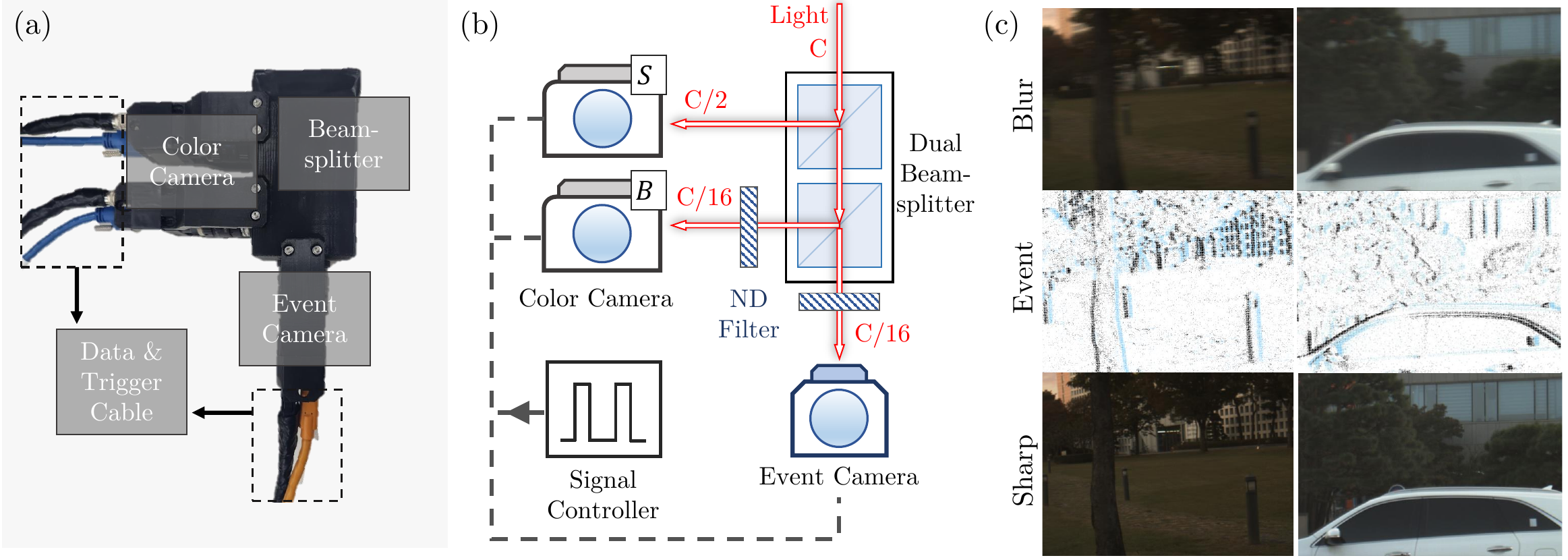}
    \caption{Illustration of a hybrid camera system for real-world event-based video deblurring dataset. $S$ and $B$ denote the cameras for acquiring sharp and blur videos, respectively.  (a): The triple-axis camera system to capture real-world blur. (b): A diagram of our hybrid camera system. (c): Samples from our EVRB dataset with natural blur.}
    \label{fig:camera_config}
\end{figure*}

\subsection{Triple-axis Hybrid Camera System}
As shown in Fig.~\ref{fig:camera_config}, we design a hybrid camera system to enable the acquisition of different data sources to be geometrically aligned.
Two RGB cameras and one event camera are geometrically aligned using two 50/50 cube beam splitters, resulting in a minimal baseline. 
We perform pixel-wise alignment between multiple cameras based on the homography calculated by extrinsic calibration for precise alignment.
For the synchronization of the multiple cameras, we use an external trigger system.
Each camera receives the falling and rising edges of the trigger signal and acquires data synchronized with the period of the signal. 
This external trigger can precisely control the exposure times of the two RGB cameras, enabling us to capture paired sharp and blurred video frames.
Furthermore, for photometric alignment, we physically adjust the amount of incoming light for both cameras to equalize the total irradiance of the multiple cameras using a neutral density filter.

\begin{figure*}[t]
\begin{center}
\includegraphics[width=1.0\linewidth]{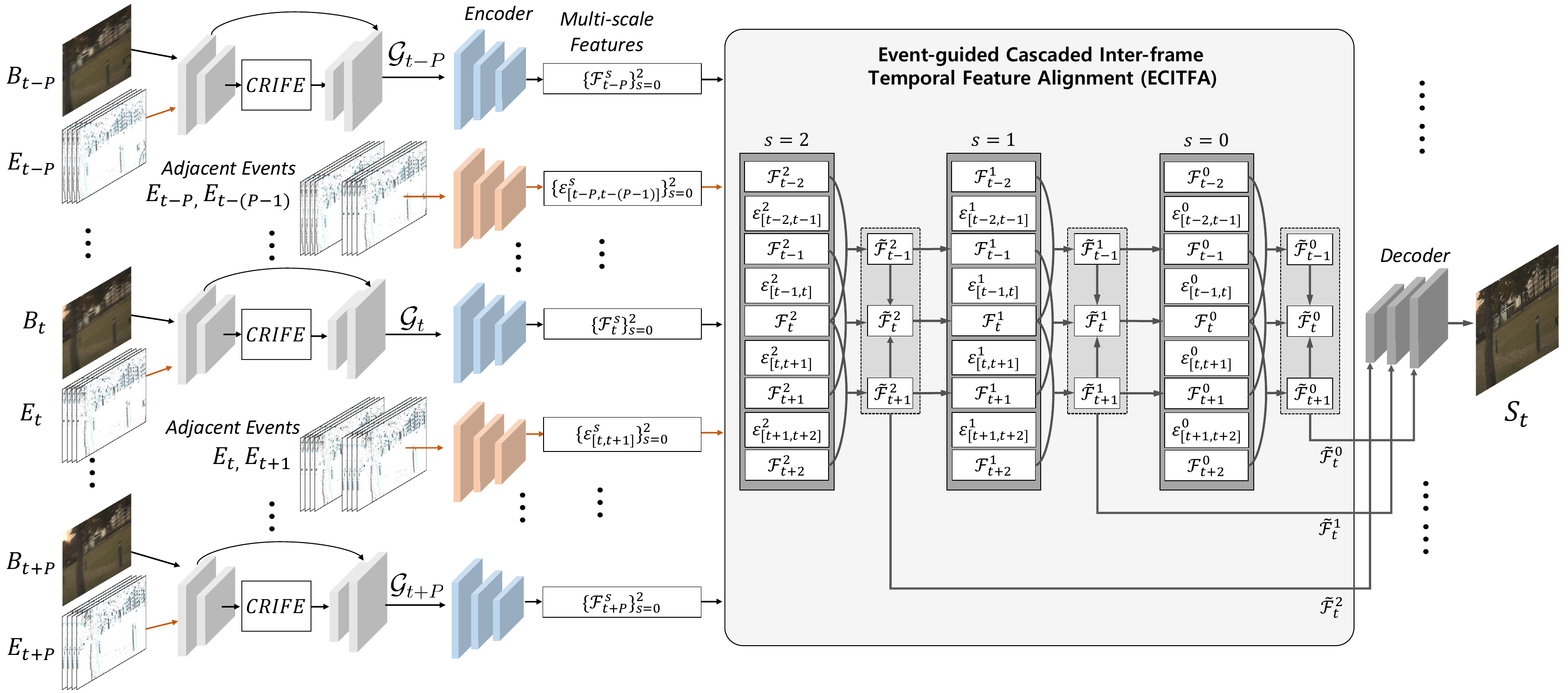}
\caption{Overall framework of CMTA is divided into two main components: Cross-modal Recurrent Intra-frame Feature Enhancement (CRIFE) and Event-guided Cascaded Inter-frame Temporal Feature Alignment (ECITFA). $s$ is the scale factor for multi-scale features.
In the figure of the ECITFA module, the description was performed for the case of $P$=2 for simplicity.
}
\label{fig:overall_framework}
\end{center}

\end{figure*}

\section{Method}
\subsection{Overview}
The overview of the proposed framework is illustrated in Fig. \ref{fig:overall_framework}.
Given consecutive blurred video frames $\{B_{k}\}$ and sets of event streams corresponding to the exposure time of each video frame $\{\mathbb{E}_{k}\}$, where $k \in\{t-P, \ldots, t, \ldots, t+P\}$, our goal is to estimate the latent sharp video frame $S_{t}$.
To utilize the event stream $\{\mathbb{E}_{k}\}$ corresponding to the exposure time of $\{B_{k}\}$ as the input for the networks, we first perform embedding using the event voxel grid representation~\cite{zhu2019unsupervised} for the event stream $\{\mathbb{E}_{k}\}$, resulting in $\{E_{k}\}$. Our framework consists of two main sub-modules: (1) Cross-modal Recurrent Intra-frame Feature Enhancement (CRIFE) module and (2) Event-guided Cascaded Inter-frame Temporal Feature Alignment (ECITFA) module.
In the first module, we perform cross-modal feature enhancement through recurrent interactions between blurred frame features and event features to leverage the continuous temporal information of events within the exposure time. After obtaining the fused feature $\{\mathcal{G}_{k}\}$ from the first module, we generate multi-scale features, $\{\mathcal{F}_{k}^s\}_{s=0}^2$, through a weight-shared pyramid encoder.
Subsequently, for the temporal feature alignment stage, we encode multi-scale event features, $\{\varepsilon_{[k, k+1]}^s\}_{s=0}^2$, by grouping two consecutive events corresponding to the exposure times of each frame to connect adjacent frames.
By utilizing the multi-scale pyramid features $\{\mathcal{F}_{k}^s\}_{s=0}^2$ and the encoded event feature $\{\varepsilon_{[k, k+1]}^s\}_{s=0}^2$, we perform temporal feature alignment. 
Afterward, temporally aligned feature pyramids are fed into the U-Net~\cite{ronneberger2015u}-based decoder, generating the final sharp video frame $S_{t}$.
Our framework can be extended to cases where $P$ is an arbitrary positive number. However, for the sake of brevity, we will explain it in the main text with $P=2$.

\begin{figure}[t]
\begin{center}
\includegraphics[width=.82\linewidth]{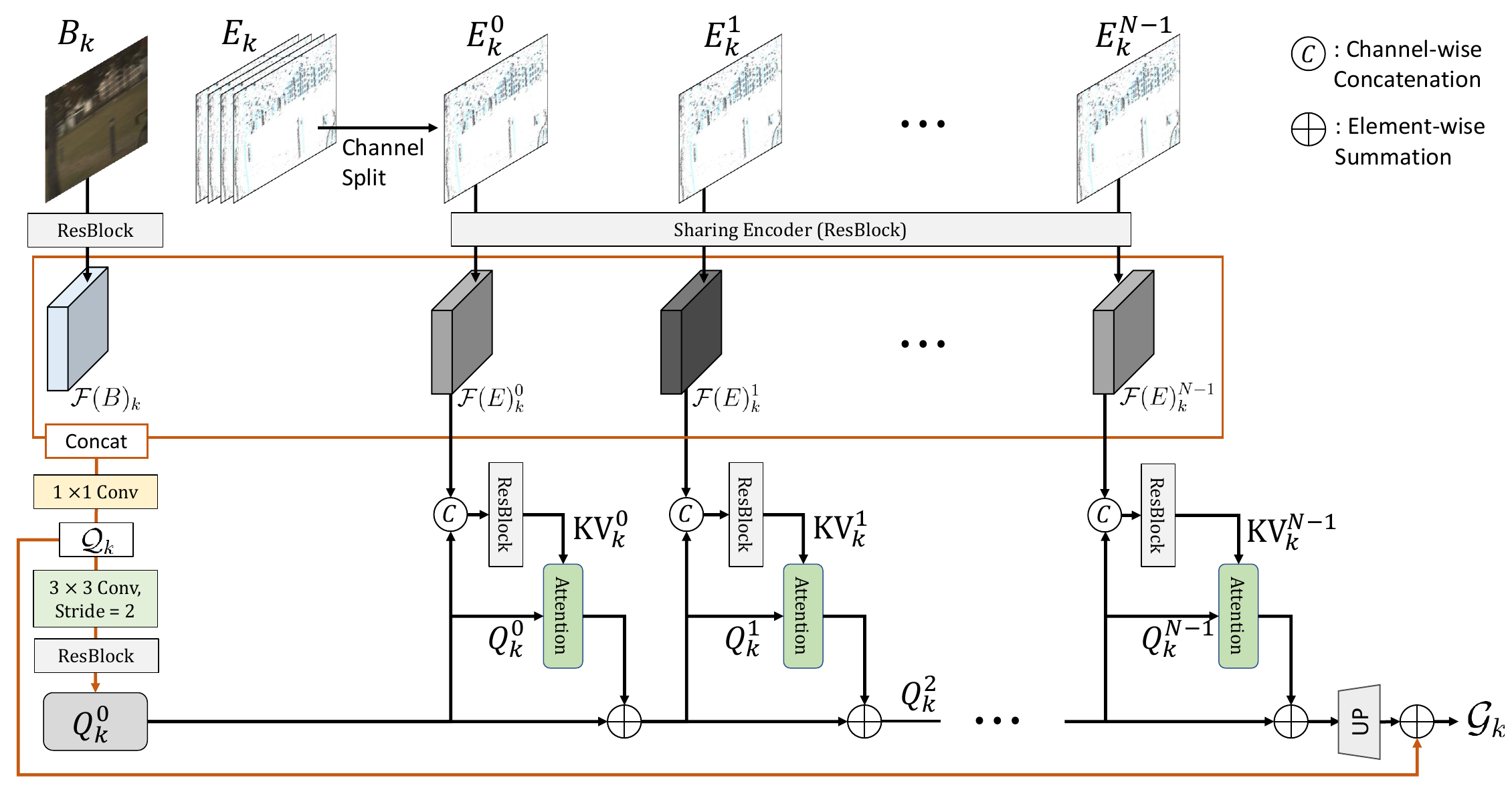}
\caption{Illustration of Cross-modal Recurrent Intra-frame Feature Enhancement (CRIFE).}
\label{fig:CRIFE}
\end{center}
\end{figure}

\subsection{Cross-modal Recurrent Intra-frame Feature Enhancement}
Since event cameras provide rich temporal information on brightness changes, it is crucial to effectively utilize this dense temporal information of the events within the duration of the exposure time.
Typically, event data captured during the exposure time is transformed into event embeddings~\cite{zhu2019unsupervised,gehrig2019end} as input to the network, followed by feature extraction using 2D CNNs.
However, these approaches cannot effectively utilize the continuous temporal information of events, and these limitations can impact the performance of event-guided video deblurring.
To address these limitations, we propose a method that leverages the rich temporal nature of events and fuses event and blurred frame features using recurrent-based attention methods.

Recently, researchers have demonstrated the effectiveness of transformer-based architectures~\cite{vaswani2017attention} by capturing long-range pixel dependencies, which have been proven highly effective in various vision tasks.  
We leverage transformers' advantages to more effectively utilize the temporal benefits of the events through a recurrent-based approach.
For each blur video frame index $k \in \{t-2, \dots, t+2\}$, we initially partition the exposure time $T_{exp,k}$ into $N$ unit temporal intervals, denoted as $\Delta t$, with $\Delta t = T_{exp,k}/N$. 
Based on the time interval $\Delta t$, we divide the event voxel grid within the exposure time, denoted as $E_{k} \in \mathbb{R}^{C \times H \times W}$ into $N$ temporally divided event voxel grids $\{E_{k}^{n}\}$, where $E_{k}^n \in \mathbb{R}^{C/N \times H \times W}$ and $n \in \{0, ..., N-1\}$.
After, we extract event features $\{\mathcal{F}(E)_{k}^{n}\}$ of the temporally divided event voxel grids $E_{k}^{n}$ using weight-sharing event feature extractor.
To apply cross-attention, we construct a query encompassing global information of blur and temporally divided event features.
That is, we concatenate the blur frame feature $\mathcal{F}(B)_{k}$ with the temporally segmented event sets $\{\mathcal{F}(E)^{n}_{k}\}$ and then extract the feature for the encoding query information.

\begin{gather}
\mathcal{Q}_k = \mathrm{Conv_{p}}(\mathcal{F}(B)_{k} \| \{\mathcal{F}(E)_{k}^{n}\}) \\
Q_k^{n=0} = F_{R}(\mathcal{Q}_k)
\end{gather}
where $\mathrm{Conv_{p}}$ represents point-wise convolution, $F_{R}$ consists of a sequence of blocks with $3 \times 3$ convolution and a stride of 2, along with $\mathrm{ResBlocks}$. Additionally, $Q_k^{n=0}$ denotes the initial query feature for the recurrent-based attention method.

As illustrated in Fig.~\ref{fig:CRIFE}, we perform cross-attention to update query information iteratively.
To find the key and value of cross-attention for query updating, we recursively input temporally separated event features, helping to better utilize the events' rich temporal information.
That is, for iteration $n$, we extract features to be used as \textit{keys} and \textit{values} by concatenating the previously updated $Q_k^{n}$ with temporally divided $n$-th event features $\mathcal{F}(E)_{k}^{n}$ as follows:
\begin{equation}
\mathrm{KV}_k^{n} = F_{KV}(Q_k^{n}\| \mathcal{F}(E)_{k}^{n})
\end{equation}
where $\|$ channel-wise concatenation and $\mathrm{KV}_k^{n}$ denotes the output features for key and value projection, and $F_{KV}$ refers to the ResBlocks layer. 
We then construct $K_k^{n}$, $V_k^{n}$ as $K_k^{n} = W^{K}(\mathrm{KV}_k^{n})$ and $V_k^{n} = W^{V}(\mathrm{KV}_k^{n})$ where $W^{(\cdot)}$ denote $1 \times 1$ convolution layer. Then, attention can be calculated as:
\begin{equation}
\mathrm{Attn_k^{n} = SoftMax}({Q_k^{n}(K_k^{n})^{T} \over \alpha})V_k^{n}
\label{equ:attention_calculation}
\end{equation}
where $\alpha$ is learnable scaling parameter to balance attention weights and $\mathrm{Attn_{n}}$ is outputs of cross-attention at iteration $n$.
To calculate cross-covariance matrix efficiently, we adopted transposed attention~\cite{zamir2022restormer} for efficient computations.
This method enables the efficient computation of attention values at high resolutions by calculating the cross-covariance matrix along the channel axis, leading to a complexity of $O(C^2)$.
Moreover, we reduced the number of channels when encoding step, enabling even more efficient operations. We use the attention value $\mathrm{Attn_{k}^{n}}$ to update the query iteratively as follows:
\begin{equation}
Q_k^{n+1} = Q_k^{n} + \mathrm{Attn_k^{n}} + \mathrm{MLP}(\mathrm{Attn_k^{n}})
\end{equation}
where $\mathrm{MLP}$ denote multi-layer perceptron.
After $N$ iterations, we obtain the updated query feature, $Q_{k}^{N}$.
The final query feature passes through up-sampling and skip connections. Then, we generate the final fused features $\{\mathcal{G}_{k}\}$ as $\mathcal{G}_{k} = \mathcal{Q}_k + \mathrm{Dconv_{4 \times 4}}(Q_{k}^{N})$ where $\mathrm{Dconv_{4 \times 4}}$ denote deconvolution layer with a kernel size of 4.
As illustrated in Fig.~\ref{fig:overall_framework}, the fused features $\{\mathcal{G}_{k}\}$ pass through a pyramid encoder, leading to the creation of multi-scale pyramid features, $\{\mathcal{F}_{k}^s\}_{s=0}^2$. Through the CRIFE module, we can leverage the advantages of abundant temporal information on the events within the exposure time.

\subsection{Event-guided Cascaded Inter-frame Temporal Feature Alignment}
Temporal feature alignment aims to extract valuable information from adjacent video frames. 
Conventional frame-based video deblurring methods face challenges from inaccurate motion estimation due to motion blur. Conversely, event cameras, thanks to their resistance to motion blur, can offer valuable guidance for aligning video frames.
The most straightforward way to align adjacent video frames using the events is to utilize event features to estimate optical flows or deformable offsets between neighboring frames~\cite{chan2021basicvsr,chan2021basicvsr++,zhu2019deformable}.
While this approach can leverage the advantages of event data for offset and optical flow estimation, it is typically employed after spatial down-sampling due to the high computational costs associated with deformable convolutions~\cite{zhu2019deformable} and optical flows~\cite{ranjan2017optical}. 
Therefore, this approach could restrict performance as it inherently hinders the processing of features at high spatial resolutions in network architectures, limiting access to information across multiple visual scales.
To address the aforementioned limitations, we propose a new temporal alignment module that combines the advantages of multiple-visual scale pyramids, leveraging the events' rich temporal contexts.

As illustrated in Fig.~\ref{fig:overall_framework}, our proposed temporal feature alignment modules are structured as multi-level networks that gradually perform coarse-to-fine feature alignment.
First, we extract event features, incorporating exposure time information from neighboring video frames to reference frames to facilitate video feature alignment.
Specifically, we encode event features encompassing the exposure time interval between $t$ and $t+1$, allowing us to connect the frame at time $t$ with the frame at time $t+1$. 
Through this event encoding step, we obtain event feature pyramid set $\{\varepsilon^{s}_{[m,m+1]}\}_{s=0}^2$ where $m\in\{t-2, ..., t+1\}$.
We use the event feature pyramid containing motion information for adjacent times for the alignment of each blur frame feature pyramid $\{\mathcal{F}_{k}^{s}\}_{s=0}^2$ where $k\in\{t-2,...,t+2\}$.

\begin{figure*}[t]
\begin{center}
\includegraphics[width=0.96\linewidth]{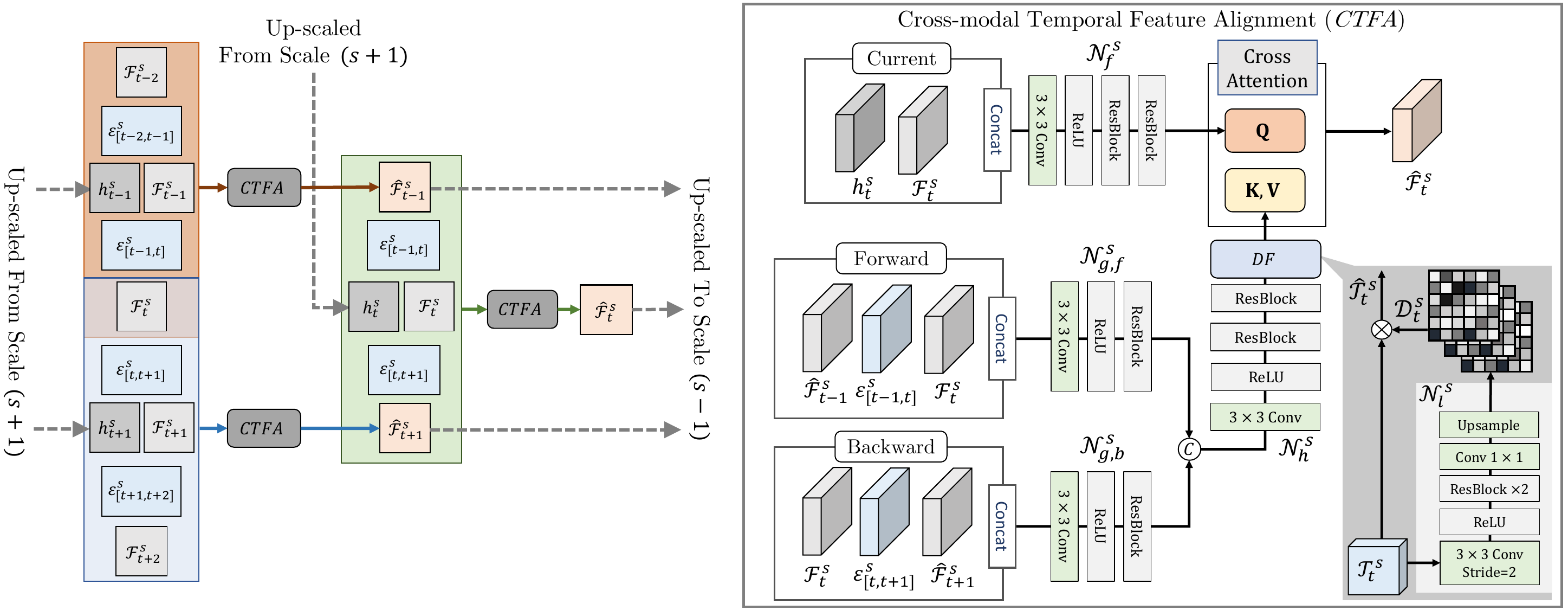}

\caption{Overview of the Event-guided Cascaded Inter-frame Temporal Feature Alignment
(ECITFA). The left figure illustrates temporal alignment for scale $s$. 
The key module for each alignment procedure, Cross-modal Temporal Feature Alignment (CTFA) at time $t$, is illustrated on the right of the figure. The CTFA module operates similarly for reference times $t-1$ and $t+1$ as well.}
\label{fig:inter_alignment}
\end{center}
\end{figure*}

After the event encoding step for the feature alignment, we gradually perform temporal feature alignment from the bottom pyramid level (scale factor $s$ of 2) to the top pyramid level ($s$ of 0).
Specifically, as shown in left side of Fig.~\ref{fig:inter_alignment}, in each pyramid level $s$ ($s \in \{0,1,2\}$), 
We first upsample the hidden state of temporally aligned features at the previous scale, $\mathcal{\hat{F}}_{i}^{s+1}$ through a deconvolution layer and receive them as inputs of alignment module, resulting in $h^{s}_{i}$.
\begin{equation}
h^{s}_{i} = \mathrm{Dconv}_{4 \times 4}(\mathcal{\hat{F}}_{i}^{s+1}), ~i \in \{t-1, t, t+1\}
\end{equation}
where $\mathcal{\hat{F}}_{i}^{s+1}$ denote aligned feature at previous scale $s+1$, $\mathrm{Dconv}_{4 \times 4}$ denote $4 \times 4$ deconvolution layer.
Note that there is no hidden state at the bottom pyramid level ($s$ of 2) since there are no features aligned at the previous scale, and for brevity, we focus on the case when $s<2$.

As depicted in the left side of Fig.~\ref{fig:inter_alignment}, we progressively group three blurred video frames when performing temporal feature alignment.
In other words, we first align with respect to $t-1$ using $\{t-2, t-1, t\}$ and simultaneously align with respect to $t+1$ using $\{t, t+1, t+2\}$ features. 
Subsequently, we perform the final temporal feature alignment for the previously aligned $t-1$ and $t+1$ with the last target time $t$.
More specifically,  when performing frame alignment for the time step $t-1$, we first group the three blur video frame features $\mathcal{F}^{s}_{t-2}, \mathcal{F}^{s}_{t-1}, \mathcal{F}^{s}_{t}$ and perform alignment. 
In this case, we make use of two event features $\varepsilon_{[t-1,t]}^s$ and $\varepsilon_{[t,t+1]}^{s}$, which respectively contain motion information between $t-1$ and $t$, and between $t$ and $t+1$, respectively.
\begin{equation}
\begin{aligned}
\hat{\mathcal{F}}_{t-1}^s= \mathrm{CTFA}^{s}(\mathcal{F}_{t}^s, \mathcal{F}_{t-1}^s, \mathcal{F}_{t-2}^s, h_{t-1}^{s}, \varepsilon_{[t-2,t-1]}^s, \varepsilon_{[t-1,t]}^s), 
\label{equ:first_align}
\end{aligned}
\end{equation}
where $\mathrm{CTFA}^{s}$ denote unit Cross-modal Temporal Feature Alignment (CTFA) module at scale factor $s$. Across the same scale factor $s$, CTFA modules are weight-shared. Similarly, we perform alignment for the time step $t+1$ as:
\begin{equation}
\begin{aligned}
\hat{\mathcal{F}}_{t+1}^s= \mathrm{CTFA}^{s}(\mathcal{F}_{t}^s, \mathcal{F}_{t+1}^s, \mathcal{F}_{i+1}^s, h_{t+1}^{s}, \varepsilon_{[t+1,t+2]}^s, \varepsilon_{[t,t+1]}^s).
\label{equ:first_align_2}
\end{aligned}
\end{equation}

Finally, we utilize the temporally aligned results $\hat{\mathcal{F}}_{t+1}^{s}$ and $\hat{\mathcal{F}}_{t-1}^{s}$ to once again group with $\mathcal{F}_{t}^s$ and perform alignment:
\begin{equation}
\begin{aligned}
\hat{\mathcal{F}}_t^s= \mathrm{CTFA}^{s}(\hat{\mathcal{F}}_{t-1}^s, \mathcal{F}_{t}^s, \hat{\mathcal{F}}_{t+1}^s,  h_{t}^{s},
\varepsilon_{[t-1,t]}^s, \varepsilon_{[t,t+1]}^s).
\label{equ:second_align}
\end{aligned}
\end{equation}

Through this cascaded feature alignment stage, we effectively propagate non-local video frame information, facilitating temporal feature alignment for both non-local and adjacent video frames.
All the results of alignment $\hat{\mathcal{F}}_{t-1}^s,\hat{\mathcal{F}}_{t}^s, \hat{\mathcal{F}}_{t+1}^s$ are passed to the next scale, $s-1$.
The right side of Fig.~\ref{fig:inter_alignment} illustrates the overall alignment process for the time step $t$ in the CTFA module.
Since it is applied similarly for all time steps $i \in \{t-1, t, t+1\}$, we will describe it here specifically when $i = t$, where previously aligned $\hat{\mathcal{F}}_{t-1}^s$ and $\hat{\mathcal{F}}_{t+1}^s$ are provided. For leveraging the benefits of coarse-to-fine architecture, we first fuse hidden state aligned feature $h_{t}^{s}$ from the previous scale for the time step $t$ at scale factor $s$, as follows:
\begin{align}
\mathcal{S}_{t}^s= \mathcal{N}_{f}^{s}(\mathcal{F}_{t}^s, h_{t}^{s}), 
\label{equ:res12}
\end{align}
We simplify the sequential operations of Conv, ReLU, and ResBlock as `$\mathcal{N}_{f}$'. 
Additionally, to leverage temporal information, we align the features at times $t-1$ and $t+1$ to the current feature at time $t$ using event features that encapsulate motion between frames in the following manner:
\begin{equation}
\begin{aligned}
&\mathcal{T}_{t-1 \rightarrow t}^s= \mathcal{N}_{g,f}^{s}(\hat{\mathcal{F}}_{t-1}^s, \mathcal{F}_{t}^s, \varepsilon_{[t-1,t]}^s), \\
&\mathcal{T}_{t+1 \rightarrow t}^s= \mathcal{N}_{g,b}^{s}(\mathcal{F}_{t}^s, \hat{\mathcal{F}}_{t+1}^s,  \varepsilon_{[t,t+1]}^s), \\
&\mathcal{T}_{t}^s = \mathcal{N}_{h}^{s}(\mathcal{T}_{t-1 \rightarrow t}^s \| \mathcal{T}_{t+1 \rightarrow t}^s),
\label{equ:res12_2}
\end{aligned}
\end{equation}

where $\|$ denotes the channel-wise concatenation, and $\mathcal{N}_{g,f}^{s}$, $\mathcal{N}_{g,b}^{s}$, and $\mathcal{N}_{h}^{s}$ represent convolution blocks, as illustrated on the right side of Fig.~\ref{fig:inter_alignment}.
While we obtain temporally aligned features $\mathcal{T}_{t}^{s}$ using event features and leveraging the advantages of events, for additional feature refinement, we utilize the spatially variant pixel-wise dynamic filter~\cite{zhou_ddf_cvpr_2021, mildenhall2018burst} mechanism.
Dynamic filter $\mathcal{D}_{t}^{s}$ at scale $s$ can be calculated through filter generation blocks, $\mathcal{N}_{l}^{s}$, which consists of convolution and resblock, such as $\mathcal{D}_{t}^{s} = \mathcal{N}_{l}^{s}(\mathcal{T}_{t}^s)$ where $\mathcal{D}_{t}^{s} \in \mathbb{R}^{(s_{k} \times s_{k}) \times H^{s} \times W^{s}}$, $s_{k}$ is kernel size of dynamic convolution filter, $H^{s}$, $W^{s}$ denote height and width of the feature at scale factor $s$, respectively.
Then, we apply the dynamic convolution operation as follows:
%
\begin{align}
\hat{\mathcal{T}}_{t}^s(h,w) = \mathcal{D}_{t}^{s}(h,w) \otimes \mathcal{T}_{t}^s(h,w)
\label{equ:dynamic_filter_oper}
\end{align}
where $h \in \{1, \dots, H^s\}, w \in \{1, \dots, W^s\}$, and $\otimes$ denotes the convolution operation.
Then, we employ a cross-attention mechanism~\cite{vaswani2017attention, zamir2022restormer} to effectively combine the information of two features, $\mathcal{S}_t^s$ and $\hat{\mathcal{T}}_t^s$.
The cross-attention generally examines the correlation between input features (query) and the key-value features.
Therefore, we conduct correlation analysis by applying cross-attention between the current blurred frame features to project query $\mathcal{S}_{t}^{s}$ and the aligned features $\mathcal{T}_{t}^{s}$ to project key and values.
That is, we generate query, key, and value features, $\mathbf{Q}= W_Q(\mathcal{S}_t^s)$,
$\mathbf{K}= W_K(\hat{\mathcal{T}}_t^s)$, $\mathbf{V}= W_V(\hat{\mathcal{T}}_t^s)$, where $W_{(\cdot)}$ is $1\times1$ convolution and $3\times3$ depth-wise convolution.
Utilizing these $\mathbf{Q}$, $\mathbf{K}$, and $\mathbf{V}$, we compute the attention matrix similarly to Eq.~(\ref{equ:attention_calculation}).
\begin{align}
\mathcal{A}(\mathbf{Q},\mathbf{K},\mathbf{V})= \operatorname{Softmax}(\frac{\mathbf{Q}\mathbf{K}^{T}}{\alpha}) \cdot \mathbf{V}
\label{equ:attention_alignment}
\end{align}
Finally, temporally aligned feature, $\hat{\mathcal{F}}_t^s$, can be obtained by:
\begin{align}
\hat{\mathcal{F}}_{t}^s = \operatorname{MLP}(\mathcal{A}) + \mathcal{A}.
\label{equ:attention}
\end{align}
where $\operatorname{MLP}$ denote multi-layer perceptrons, $\mathcal{A}$ denote the result of attention operation.
Finally, we obtained the aligned feature pyramid $\{\hat{\mathcal{F}_{t}^{s}}\}$ through cascaded temporal feature alignment.
As illustrated in Fig.~\ref{fig:overall_framework}, these multi-scale aligned features are fed into the decoder.

\subsection{Decoder}
The decoder is designed based on the standard U-Net~\cite{ronneberger2015u}. 
It takes multi-scale temporally aligned features $\{\hat{\mathcal{F}}_{t}^{s}\}$ as inputs and produces output feature pyramid $\{\mathcal{F}(D)_{k}^{s}\}$.
The final deblurred outputs $S_{t}$ using last scale of output feature $\mathcal{F}(D)_{t}^{s=0}$ is calculated as follows:
\begin{equation}
\begin{aligned}
S_{t} = B_{t} + \mathrm{Conv_{5x5}}(\mathcal{F}(D)_{t}^{s=0})
\end{aligned}
\end{equation}
where $\mathrm{Conv_{5x5}}$ represents a conv. layer with a filter size of $5\times5$, and $S_{t}$ denotes the final estimated sharp frame.

\section{Experiments}

\begin{table}[!t]
\centering
\caption{Quantitative results on the GoPro dataset. The asterisk($*$) indicates that the results are not officially on the GoPro so we retrained the official model by us. CMTA-5 and CMTA-7 refer to the results obtained using 5 and 7 input frames, respectively.}
\resizebox{1.00\linewidth}{!}{
\begin{tabular}{c|ccccccccc}
\hline
Methods & MPRNet~\cite{zamir2021multi} &  Restormer~\cite{zamir2022restormer} & NAFNet~\cite{chen2022simple} & EDVR~\cite{Wang_2019_CVPR_Workshops}  & ESTRNN~\cite{zhong2020efficient} & RNN-MBP~\cite{zhu2022deep} & DSTNet~\cite{Pan_2023_CVPR} & VRT~\cite{liang2022vrt} \\ \hline
PSNRs & 32.66 &  32.92 & 33.69  & 26.83 &  31.02 & 33.32 & 34.16 &  34.81 \\
SSIMs & 0.959  & 0.961 & 0.967  & 0.843  & 0.911 & 0.963 & 0.968 & 0.972\\
Params(MB) & 20.1 & 26.1 & 67.8  & 23.6  & 2.4 & 16.4 & 7.5 & 18.3 \\ \hline
Methods  & RVRT~\cite{liang2022recurrent} & Shift-Net~\cite{Li_2023_CVPR}  & UEVD$^{*}$~\cite{kim2022event} & EFNet~\cite{sun2022event} & REFID~\cite{sun2023event} & SpkNet~\cite{chen2024enhancing} & \textbf{CMTA-5} & \textbf{CMTA-7} \\ \hline
PSNRs & 34.92 & 35.49 & 35.48 & 35.46 & 35.91 & 36.12 & 36.55  & \textbf{36.78} \\
SSIMs   & 0.974 & 0.976  & 0.971 & 0.972& 0.973 & \multicolumn{1}{c}{0.971} & \textbf{0.977} & \textbf{0.977} \\
Params(MB) & 10.8 & 10.5 & 27.9 & 8.5 &  15.9 & \multicolumn{1}{c}{13.5} & 9.7 & 9.7 \\ \hline
\end{tabular}
}
\centering
\label{tab:gopro}
\end{table}

\begin{figure*}[t]
\centering\includegraphics[width=1\linewidth]{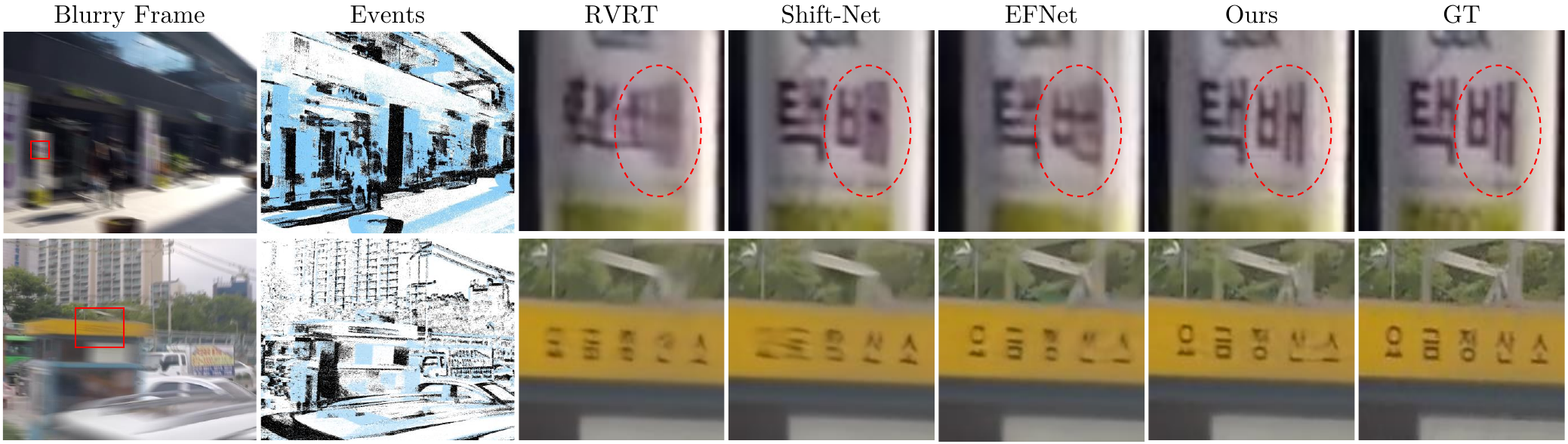}
    
    \caption{Visual comparison of deblurring results on the GoPro dataset. The qualitative results of other methods were taken from the results provided by the authors.}
    \label{fig:main_qual_gopro}
\end{figure*}

\subsection{Datasets}
\noindent
\textbf{GoPro Dataset~\cite{nah2017deep}.}
For a fair comparison with other previous event-guided deblurring methods, we utilize the same raw events provided by the authors of recent work~\cite{sun2022event}.
These events were generated using ESIM~\cite{rebecq2018esim} with a randomly generated contrast threshold set to a Gaussian  normal distribution of parameters as $N(\mu = 0.2, \sigma = 0.03)$. 
We used the official train and test splits.

\noindent
\textbf{HighREV Dataset~\cite{sun2023event}.}
HighREV consists of high-resolution events and RGB data at $1632\times1224$ resolution, designed for both deblurring and interpolation tasks. To evaluate motion deblurring exclusively, we use the \textit{11}+\textit{1} split, excluding the interpolation ground truth.

\noindent
\textbf{Real-world Video Deblurring Dataset.}
EVRB dataset consists of 11 training sequences and 6 test sequences. 
With each sequence containing 149 frames, it is well-suited for video deblurring tasks. 

\begin{table}[!t]
\centering
\caption{Quantitative results on the HighREV dataset.}
\resizebox{.95\textwidth}{!}{
\setlength\tabcolsep{10.0pt}
\begin{tabular}{c|cc|cccc}
\hline
Methods & ESTRNN~\cite{zhong2020efficient} & DSTNet~\cite{Pan_2023_CVPR} & UEVD~\cite{kim2022event} & EFNet~\cite{sun2022event} & REFID~\cite{sun2023event} & Ours \\ \hline
PSNRs & 30.38 & 31.77 & 37.40 & 37.99 & 38.37 & \textbf{39.12} \\
SSIMs & 0.940 & 0.948 & 0.974 & 0.976 & 0.977 & \textbf{0.980} \\ \hline
\end{tabular}
}
\centering
\label{tab:highrev}
\end{table}

\subsection{Comparison on Synthetic Blur Datasets}
We present the quantitative results of our frameworks with other frame-based image and video deblurring methods and event-guided motion deblurring methods using the GoPro dataset as depicted in Table~\ref{tab:gopro}.
When compared to the existing best performance of video deblurring methods, Shift-Net~\cite{Li_2023_CVPR}, our approach (CMTA-5 model) shows a significant improvement of 1.06 dB in terms of PSNR, demonstrating that our method effectively leverages the temporal dense characteristics of event modality for accurate temporal feature alignment and cross-modality feature enhancement. 
Moreover, compared to the best-performing SpkNet~\cite{chen2024enhancing} among existing event-guided motion deblurring methods, CMTA-5 model exhibits a performance improvement of 0.43 dB with a lower model params of 3.8 MB.
Furthermore, by using 7 input video frames (CMTA-7 model) instead of 5 (CMTA-5 model), we achieved an impressive state-of-the-art performance with a PSNR of 36.78dB in the GoPro, an increase of 0.23.
We further demonstrated the superiority of our approach through the qualitative results in Fig.~\ref{fig:main_qual_gopro}.
Also, we conduct experiments on HighREV~\cite{sun2023event}, which consists of real events, and report the results in Table~\ref{tab:highrev}. Our method still achieves the best performance.

\subsection{Comparison on Real-world Blur Datasets}
\begin{figure*}[t]
    \centering
    \includegraphics[width=1\linewidth]{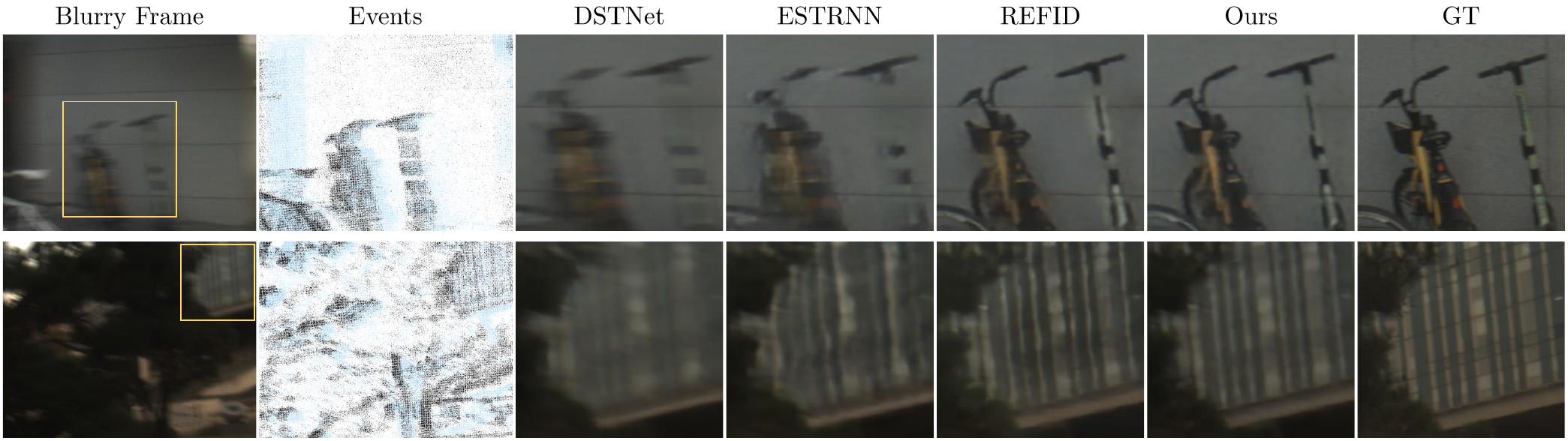}
    \caption{Visual comparison of deblurring results on the EVRB dataset.}
    \label{fig:main_qual}
\end{figure*}

For comparisons in the EVRB dataset, we trained representative video deblurring methods~\cite{Wang_2019_CVPR_Workshops,zhong2020efficient, Pan_2023_CVPR,jiang2022erdn,Li_2023_CVPR, liang2022recurrent} and event-guided motion deblurring methods~\cite{sun2022event,sun2023event,kim2022event} on the same training set.
Tab~\ref{tab:evrb} presents the quantitative results on the EVRB dataset. 
The EVRB dataset includes extremely blurred videos captured during exposure times.
As a result, frame-based video deblurring methods exhibit subpar performance. 
For instance, the best-performing network, Shift-Net~\cite{Li_2023_CVPR}, achieves only 30.56 dB, which is 0.42 dB lower than the best-performing event-guided deblurring method, EFNet~\cite{sun2022event}. 
In contrast, our approach leverages video and event characteristics, effectively restoring even severe motion blur, which may be difficult to recover.
Our method outperforms all approaches, achieving the best performance.
We show our qualitative comparison with other methods in Fig.~\ref{fig:main_qual}.

\newcolumntype{P}[1]{>{\centering\arraybackslash}p{#1}}

\begin{table}[!t]
\centering
\caption{Quantitative results on the EVRB dataset.}
\resizebox{.96\textwidth}{!}{
\setlength\tabcolsep{12.0pt}
\begin{tabular}{c|cccccccccc}
\hline
Methods & EDVR~\cite{Wang_2019_CVPR_Workshops} & ESTRNN~\cite{zhong2020efficient} & ERDN~\cite{jiang2022erdn} & DSTNet~\cite{Pan_2023_CVPR} & RVRT~\cite{liang2022recurrent} \\ \hline
PSNRs & 29.02 & 29.79 & 28.32 & 29.15 & 30.24 \\
SSIMs & 0.886 & 0.911 & 0.893 & 0.898 & 0.906 \\ 
\hline
Methods &  Shift-Net~\cite{Li_2023_CVPR} & UEVD~\cite{kim2022event} & EFNet~\cite{sun2022event} & REFID~\cite{sun2023event} & Ours \\ \hline
PSNRs & 30.56 & 30.55 & 30.98 & 30.33 & \textbf{31.38} \\
SSIMs & 0.922 & 0.915 & \textbf{0.927} & 0.918 & \textbf{0.927} \\ \hline
\end{tabular}
}
\centering
\label{tab:evrb}
\end{table}

\subsection{Ablation Study}
We analyzed the performance contribution of the various modules in our frameworks. For a fair ablation study, we trained all the models for 600 epochs with 5 video frame inputs, conducting all experiments on the GoPro dataset.

\noindent
\textbf{CRIFE module.} 
To demonstrate the effectiveness of the CRIFE module, we replaced it with concatenation and convolution for comparison.
When comparing the first column(Ver.1) and the second column(Ver.2) of Tab ~\ref{tab:ablation_gopro}, we observed a performance gain of +0.35 dB with a small additional parameter (+0.08 MB).
Similarly, when comparing the performance of the third and fourth rows, we observed performance improvement.
\\
\noindent\textbf{ECITFA module} is the most crucial component of our model, performing feature alignment by leveraging information of the events.
As in the Tab.~\ref{tab:ablation_gopro}, when comparing the first (Ver.1) with the third column (Ver.3) of the table, we observed a significant performance gain (\textbf{+1.69} dB) upon the insertion of the ECITFA module.
This trend is similarly maintained when the CRIFE module is incorporated.
When comparing Ver.2 with Ver.4 of the Tab.~\ref{tab:ablation_gopro}, we observed a notable performance improvement (\textbf{+1.55} dB).

\noindent\textbf{Effectiveness of components in ECITFA module.} 
In the Tab.~\ref{tab:ECITFA}, we demonstrated an effectiveness analysis for each component of ECITFA.
Comparing the 2nd column with the last column labeled `Ours', we observed a performance gain of our method (+0.37 dB) when employing spatial pixel-wise dynamic filters, in contrast to not using $\mathrm{DF}$.
When utilizing cross-attention (Eq.\ref{equ:attention_alignment}) to better leverage long-range pixel-dependencies in the alignment blocks, we observed a performance improvement (+0.29 dB) compared to aggregate feature using $\mathrm{ResBlocks}$.
Finally, we confirmed the effectiveness of leveraging non-local video frame information through a cascaded-based temporal feature alignment method for video deblurring.
After using the proposed cascaded structure for temporal feature alignment, we can observe a performance gain (+0.47 dB).

\begin{table}[!t]
\centering
\caption{Ablation study on the GoPro dataset.}

\setlength\tabcolsep{6.5pt}
\resizebox{.85\textwidth}{!}{
\begin{tabular}{c|c|c|c|c}
\hline
Methods & Ver.1 & Ver.2 & Ver.3 & Ver.4 \\ \hline
CRIFE &   & \checkmark  & & \checkmark  \\
ECITFA & &  & \checkmark & \checkmark  \\ \hline
PSNRs / Params & 34.65 / 4.57M & 35.00 / 4.65M & \underline{36.34} / 9.59M & \textbf{36.55} / 9.68M \\
\hline
\end{tabular}
}

\centering
\label{tab:ablation_gopro}
\end{table}

\noindent\textbf{Effectiveness of components in CRIFE module.} 
To evaluate each component's effectiveness of the CRIFE module, we removed the concatenation between RGB and event features, directly matching their features. However, as shown in the 3rd column of Tab.~\ref{tab:fusion_varaint_module}, this approach yielded sub-optimal performance. Additionally, combining event features without a recurrent structure degraded performance (see 4th column). These ablations confirm that the recurrent structure effectively utilizes temporal information from events within exposure time.


\begin{table}[!t]
\centering
\caption{Effect of the component of ECITFA module on the GoPro dataset. $\mathrm{DF}$ denotes dynamic filter operation in the Eq.(\ref{equ:dynamic_filter_oper}).
}

\setlength\tabcolsep{10.5pt}
\resizebox{.85\textwidth}{!}{
\begin{tabular}{c|ccc|c}
\hline
Methods & w/o $\mathrm{Cascaded}$ & w/o $\mathrm{DF}$ & w/o $\mathrm{Attention}$ & Ours \\ \hline
PSNRs & 36.08 & 36.18 & \underline{36.29} & \textbf{36.55} \\ \hline
\end{tabular}
}

\centering
\label{tab:ECITFA}
\end{table}

\begin{table}[!t]
\centering
\caption{Comparison of CRIFE with various module variants. ``Baseline'' refers to Ver.3 of Tab.\ref{tab:ablation_gopro}. E and F represent event and frame features, respectively.}
\centering
\setlength\tabcolsep{8pt}
\resizebox{.85\textwidth}{!}{
\begin{tabular}{c|c|c|cc}
\hline
\multirow{2}{*}{Methods} & \multirow{2}{*}{Baseline} & \multirow{2}{*}{w/o concat (E+F)} & \multicolumn{2}{c}{w/concat (E+F)} \\ \cline{4-5} 
 &  &  & \multicolumn{1}{c|}{w/o recurrent} & w/ recurrent(Ours) \\ \hline
PSNRs & \underline{36.34} & 36.04 & \multicolumn{1}{c|}{35.99} & \textbf{36.55} \\ \hline
\end{tabular}
}
\label{tab:fusion_varaint_module}
\centering
\end{table}

\section{Conclusions}
This paper proposes a video deblurring framework, CMTA, that elaborately considers the characteristics of an event and video.
Specifically, we achieve significant performance improvement through intra-frame feature enhancement and inter-frame temporal feature alignment.
Furthermore, we construct a real-world deblurring dataset, the EVRB dataset, which will be valuable for evaluating event-guided deblurring methods.
Finally, CMTA demonstrates state-of-the-art performance across various deblurring datasets.

\noindent
\textbf{Acknowledgements.} This work was supported by the Technology Innovation Program (1415187329,20024355, Development of autonomous driving connectivity technology based on sensor-infrastructure cooperation) funded By the Ministry of Trade, Industry \& Energy(MOTIE, Korea) and the National Research Foundation of Korea(NRF) grant funded by the Korea government(MSIT) (NRF2022R1A2B5B03002636).

%
%
\bibliographystyle{splncs04}

\end{document}